# IMAGE-BASED AGARWOOD RESINOUS AREA SEGMENTATION USING DEEP LEARNING

Irwandi Hipiny        Johari Abdullah        Noor Alamshah Bolhassan

**Abstract**

The manual extraction method of Agarwood resinous compound is laborious work, requires skilled workers, and is subject to human errors. Commercial Agarwood industries have been actively exploring using Computer Numerical Control (CNC) machines to replace human effort for this particular task. The CNC machine accepts a G-code script produced from a binary image in which the wood region that needs to be chiselled off is marked with (0, 0, 0) as its RGB value. Rather than requiring a human expert to perform the region marking, we propose using a Deep learning image segmentation method instead. Our setup involves a camera that captures the cross-section image and then passes the image file to a computer. The computer performs the automated image segmentation and feeds the CNC machine with a G-code script. In this article, we report the initial segmentation results achieved using a state-of-the-art Deep learning segmentation method and discuss potential improvements to refine the segmentation accuracy.

## 1 Introduction

Agarwood, also known as "oud" or "gaharu," is a highly prized aromatic resin derived from the Aquilaria species of trees. When Aquilaria trees undergo a natural or induced process of resin formation, the tree's inner core undergoes chemical changes and produces a resinous compound. The appearance of the compound is characterised by its dark and dense nature, exhibiting a range of colours from deep brown to black. Over time, the resin accumulates and solidifies, producing a hardened compound that releases an aromatic scent when burned. The hardened compound is collected by scraping or chiselling it off the wood using a specialised hand tool. The laborious work requires skilled workers, is time-consuming, and is subject to human errors that lead to waste. Considering the high market value of Agarwood resin, this potential wastage in terms of production time and output is not ideal.

Commercial Agarwood industries have been actively exploring using Computer Numerical Control (CNC) machines to extract the resin. The goal is to replace the human worker with a fully automated machine. The CNC machine accepts a G-code script as input. G-code refers to a programming language consisting of a series of commands, or instructions, sent to the CNC machine to execute various aspects of the machining process, such as tool movement, spindle speed, feed rate, tool changes, and coolant activation. Generating G-code for CNC machines requires a preliminary step of obtaining a segmentation result or a digital representation of the desired object or part. The segmentation process involves marking individual image regions as either background or foreground. This step can be done manually using image editing software or in an automated manner via image segmentation.

The main problem with using such algorithms is the accuracy of the produced segmentation. The algorithm must be able to differentiate between the resinous compound (foreground) and the rest of the wood surface (background). Our work proposes using a state-of-the-art Deep learning segmentation algorithm, i.e., Segment Anything Model (SAM), by Kirillov et al. (2023). The model is prompt-able, meaning that the initial segmentation result can be further refined with manual prompts to mark a region's membership. A complete autonomous workflow is possible by automating the prompts. In this work, we report our initial results using Kirillov et al. (2023) SAM model on a small dataset containing cross-section images of Agarwood trees. A visualisation of the proposed workflow is presented in Figure 1 (a).

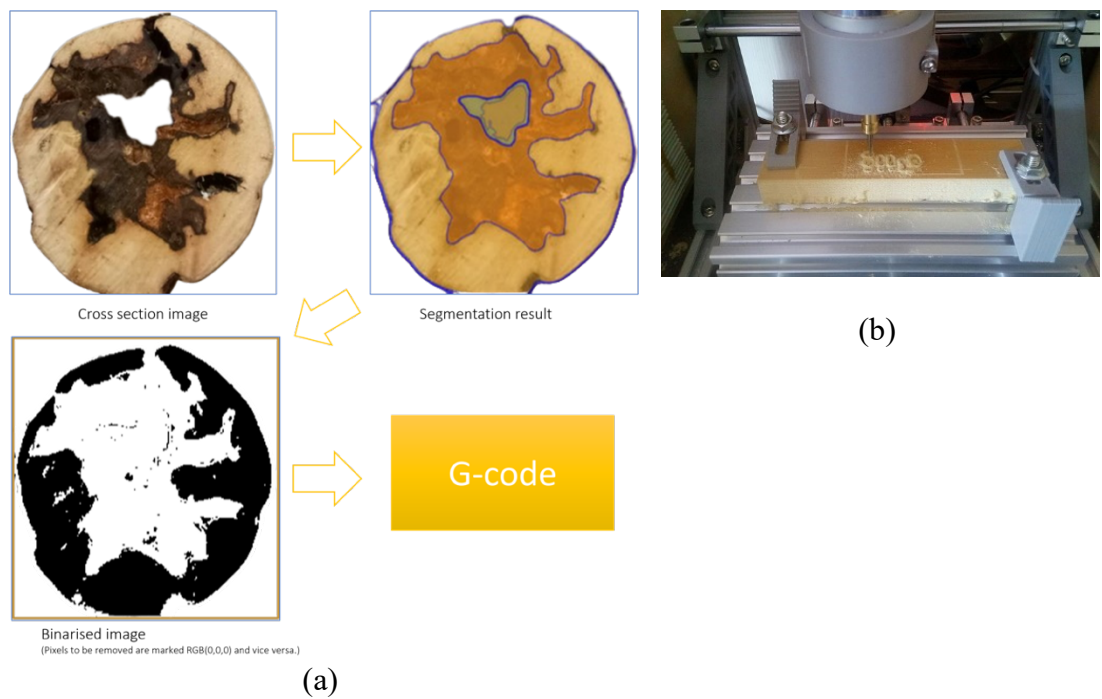

(a)

Fig.1 a) A visualisation of our proposed workflow. Given a cross-section image (with the background removed), we use Kirillov et al. (2023) SAM model to segment the possible region(s) inside. At this stage of our work, we do not refine the segmentation result using additional prompts. The segmented image is then binarised before being converted to a G-code script. b) A mini CNC engraver machine. The image is sourced from Fixerdave (2018).

The rest of the article is organised as follows: In Section 2, we reviewed Deep learning image segmentation methods. We only focused on methods used in wood cross-section images. Section 3 describes our dataset and then outlines our method. We also explain the metric used to rate the segmentation results. In Section 4, we reported and discussed our experimental results. Finally, we conclude our work and offer potential future works in Section 5.

## 2 Previous Work

Several works exist in the literature on segmenting region(s) of interest inside a wood cross-section image using Deep learning. Decelle and Jalilian (2020) compared several deep Convolutional Neural Network (CNN) based segmentation methods, i.e., U-Net (Ronneberger et al., 2015), Mask R-CNN (He et al., 2017), RefineNet (Lin et al., 2017) and SegNet (Badrinarayanan et al., 2017) of different-sized wood cross-section (CS) RGB image datasets. The comparison results show that U-NET outperforms the others on smaller datasets while RefineNet learns well on larger datasets. Yoo et al. (2022) segment rays in tangential thin sections of conifers using Mask R-CNN (He et al., 2017). They achieved a mean average precision of 0.837; however, Mask R-CNN tends to produce over-segmented regions. Gillert et al. (2023) propose a novel iterative segmentation method., i.e., Iterative Next Boundary Detection (INBD), to segment tree rings. Given a microscopy image of shrub cross-sections, the method starts from the centre and detects the next ring boundary iteratively. INDB reports, on average, a higher mean Average Recall than the other state-of-the-art segmentation methods, including Mask R-CNN (He et al., 2017) and Multicut (Kappes et al., 2017).

Meta AI's Segment Anything Model (Kirillov et al., 2023) is a Transformer-based vision model containing the following components, i.e., an image encoder, a flexible prompt encoder and a fast mask decoder, see Figure 2. The model was trained on a custom dataset containing over a billion masks on 11 million images.

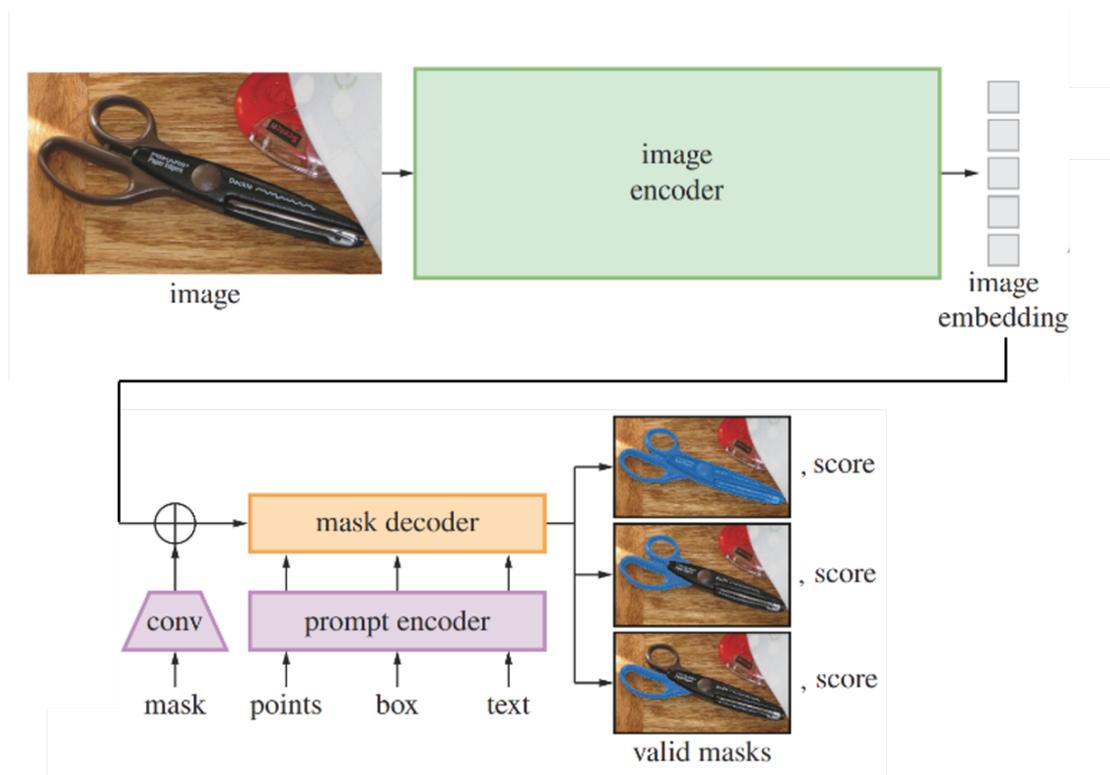

Fig. 2 An overview of Kirillov et al. (2023) Segment Anything Model. The model accepts an RGB image and encodes the image embeddings before passing them to a conv mask. The model then accepts three prompt types, i.e., points, box and text, as input for the mask decoder to produce region proposals with a confidence score each. The above image is sourced from Kirillov et al. (2023).

SAM produces region proposals sorted according to their confidence scores. These region proposals are generated by a mask decoder that accepts multimodal prompts in the form of points, boxes and texts. The model allows zero-shot transfer to new image distributions and tasks, allowing it to work with never-seen-before images. Also, SAM is prompt-able; by default, the model is prompted with a 16×16 regular grid of foreground points. Redundant prompts are removed based on their visual similarity, and the remaining ones are used to produce the region proposals. The segmentation result can be further refined using additional custom prompts. Carefully chosen prompts will improve the segmentation result, whilst poor ones will decrease the accuracy.

To the best of our knowledge, only a single work in the literature, i.e., Ishak et al. (2017), performs segmentation on Agarwood cross-section images. Instead of using a Deep learning segmentation method, Ishak et al. (2017) used simple binary thresholding to segment the resinous regions inside a Greyscale image. The segmented regions are then referred to generate a G-code script for the CNC machine. However, they did not report the segmentation accuracy, as the main goal of their work is to classify agarwood grading. Ishak et al. (2017) used a grading system shown in Table 1. The resin has varying colours and hues that determine the grade.

Table 1: Agarwood Grading System (*Manual Penggredan Gaharu, 2015*).

| Grade | Resin Colour |
|---|---|
| Super A | Combination of all possible colours, with an attractive shape |
| A | Black or Shiny Black |
| B | Brown or Dark Brown |
| C | Whitish or Yellowish |

## 3 Method and Dataset

### 3.1 Dataset

Our small dataset is acquired from the personal collection of our sole contributor. The cross-section images were produced by cropping the two source images shown in Figure 3, resulting in 12 smaller images. The cropped images were not subjected to any additional pre-processing. At this stage of our work, we decided against using any image filters to enhance (or subdue) the appearance of specific image features inside the cross-section image. This decision was made because we wish to evaluate the chosen segmentation algorithm by its own merit.

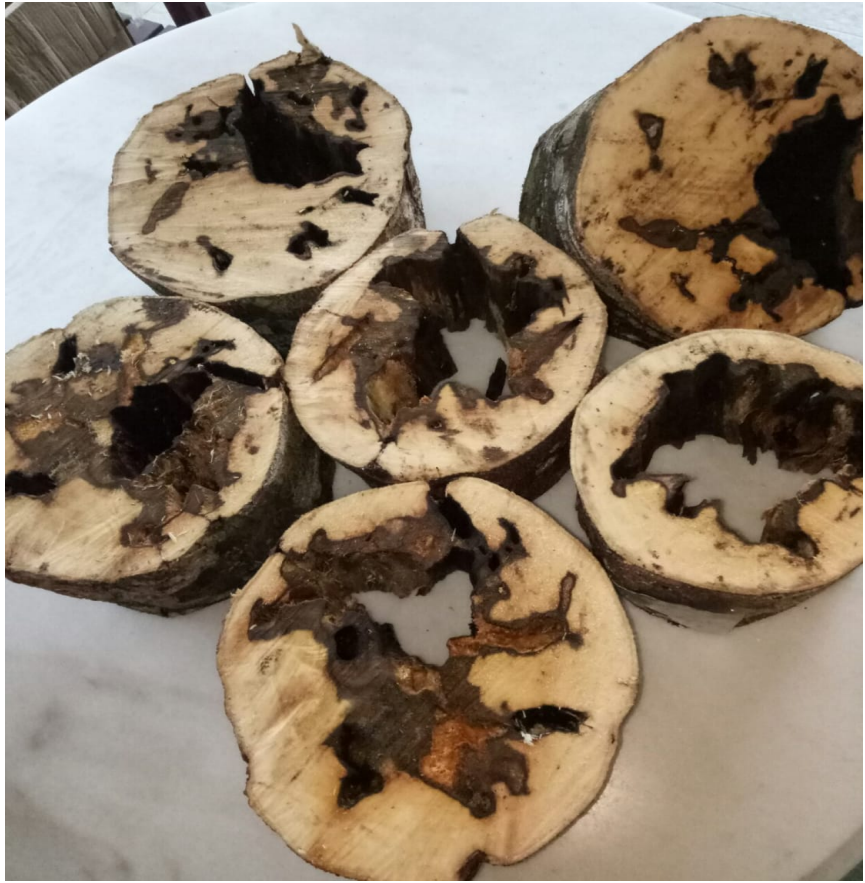

(a)

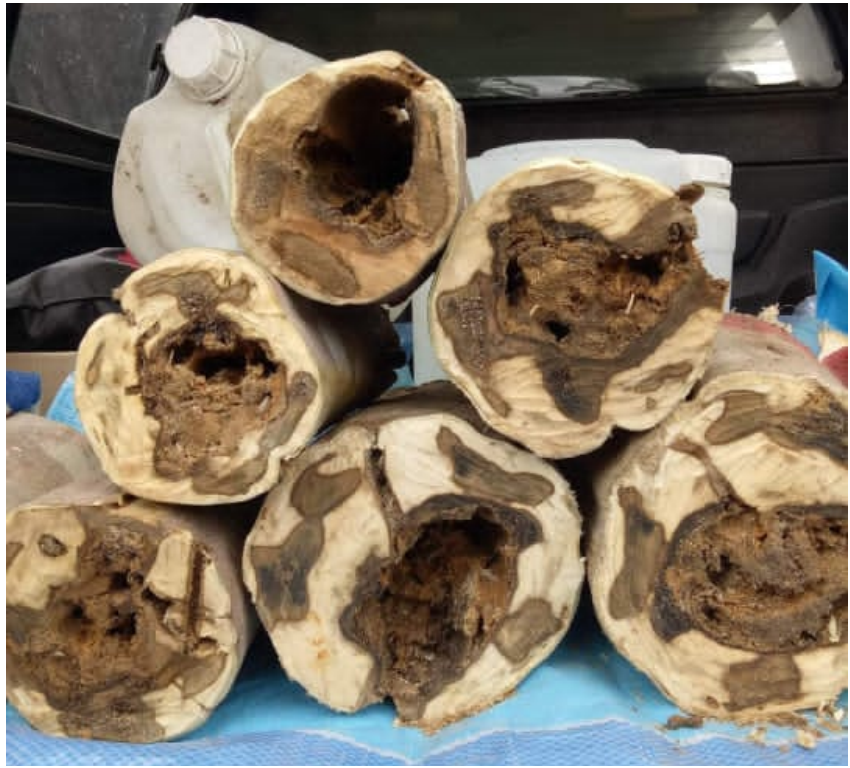

(b)

Fig. 3 Agarwood wood slices. Both images were contributed by Bakar (2023).

As can be observed from the 12 cropped images, the cross-section wood area can be divided into three categories, i.e., the healthy region, the resinous compound region, and the decayed core region (caused by heart rot). The healthy region is markedly different from the other two regions, as the colours are considerably lighter. The resinous compound and the decayed core share visual similarities (i.e., darker colours). During manual extraction, the worker differentiates between the two regions based on physical texture. The resin compound has a hardened texture, while the decayed core region is much softer to the touch. See Figure 4 for a visual explanation of the region classifications.

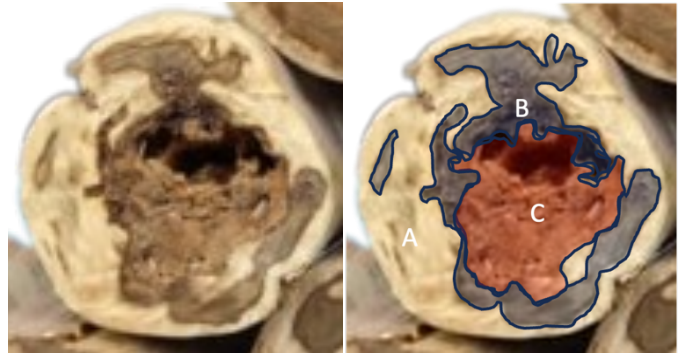

Fig. 4 Visual explanation of the region classifications, i.e., A) Healthy region, B) Resinous compound, and (c) Decayed core region.

At this stage of our work, we do not implement any additional processing to differentiate between the two regions. Our current implementation only produces a binary classification, i.e., Removed region (lighter coloured area) vs Retained region (darker coloured area). We manually marked the Retained region for each image to produce our ground truth data. Later, we measure the overlap between the segmented regions and the ground truth data to obtain the segmentation accuracy. The images and the corresponding ground truth masks are stored in RGB colour space and PNG file format and available for download (CC BY-NC 4.0) at the author's website.

## 3.2 Method

Our proposed method employs Kirillov et al. (2023) Segment Anything Model as the Deep learning segmentation method. Figure 5 below shows a pseudocode of the proposed method's algorithm. We first perform a background removal on an input RGB image. This is a standard function available in most Computer vision libraries. The function estimates the foreground mask by performing a subtraction between the current frame and a background model. By placing the wood slice on a uniformly coloured flat surface, the background region can be easily segmented and removed. We suggest using a matte work surface with a highly contrasting surface colour, compared to the wood slice's colours, such as green or blue.

| ALGORITHM 1 |
|---|
| set i = 0 |
| read input image |
| estimate foreground mask |
| create a 16×16 regular grid of prompts |
| add prompt$_{i'th}$ into KEEP |
| while i < (16×16), do |
|     if dist(prompt$_{i'th}$, KEEP) > thresh, then add prompt$_{i'th}$ into KEEP |
|     i++ |
| end while |
| produce region proposals |

Fig. 5 Pseudocode.

The resulting image is then passed to Kirillov et al. (2023) Segment Anything Model (SAM) to produce the region proposals, each with a confidence score ranked from highest to lowest. By default, SAM employs a 16×16 regular grid of prompts from which redundant prompts are removed. Prompts that are visually similar are removed, with the centroid retained as the group's single representative. Only the remaining prompts are used to produce the region proposals. Next, the segmented image is converted to a binary image where each pixel has an RGB value of (0, 0, 0) or (255, 255, 255). RGB (0, 0, 0) pixels are marked for removal and vice versa. Finally, the binary image is converted to a G-code script that a CNC machine can read. Our work used an online free converter (Huijzer, 2013) available at the following URL: https://thuijzer.nl/image2gcode/.

To evaluate the segmentation results produced by SAM, we measure the Intersection over Union (IoU),

$$IoU = \frac{Area\ of\ Overlap}{Area\ of\ Union} \qquad (1.0)$$

between the predicted vs the actual Retained region. The value is normalised in a range between 0.0 and 1.0, where a higher value indicates a more accurate segmentation (i.e., a greater overlap). This metric has been used in the literature, e.g., by John et al. (2017), to indicate the overlap between two individual image regions.

## 4 Results & Discussion

We present the default SAM-produced segmentation result for each cropped image in Figure 6. We include the "before" and "after" images side-by-side to ease segmentation quality assessment based on visual inspection. We then compute the Intersection over Union (IoU) for each image and tabulate the result in Table 2.

| **Cropped Image** (w/ background removed) | **Segmentation result** (using default SAM) | **Seg. Qua.** |

(a) 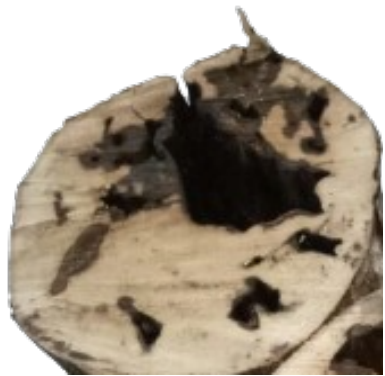 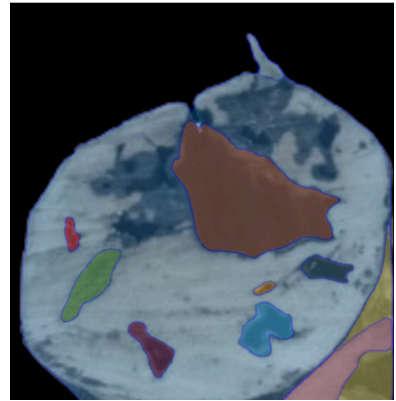

Mod.

(b) 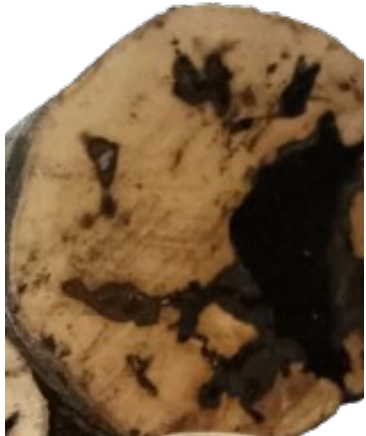 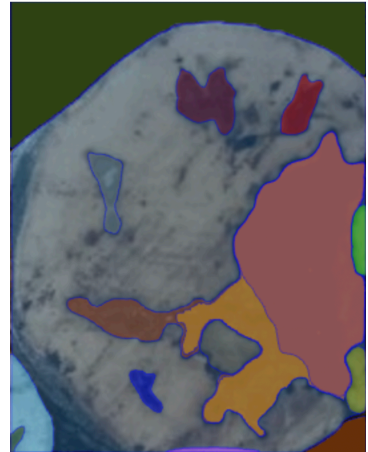

Good

(c) 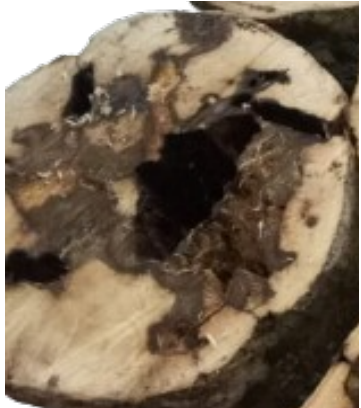 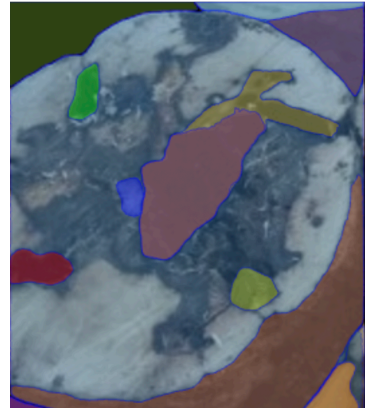

Poor

(d) 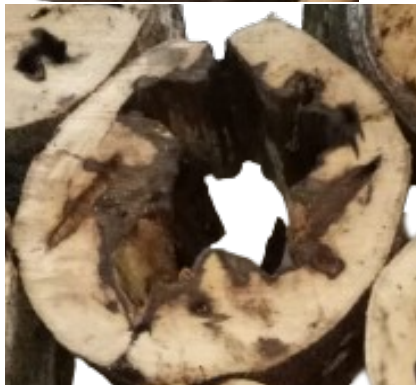 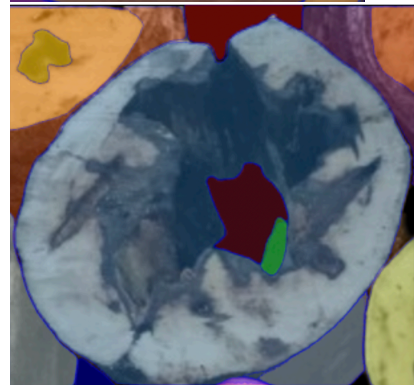

Poor

(e) 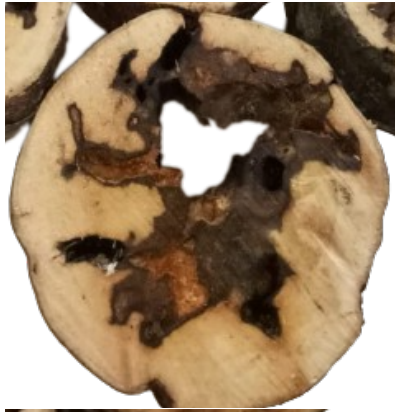 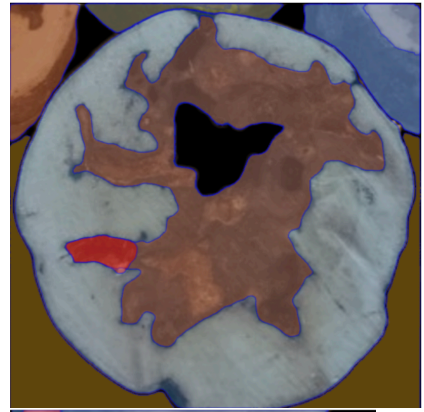 Good

(f) 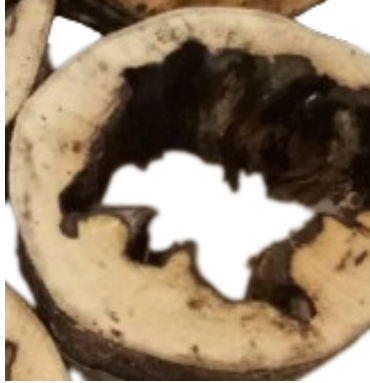 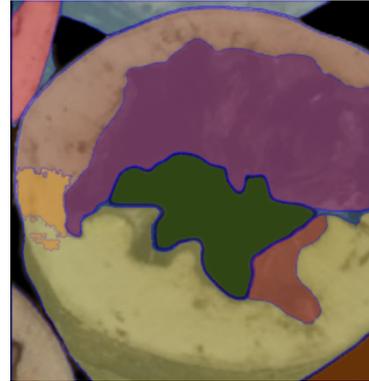 Good

(g) 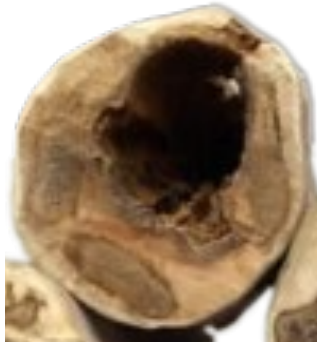 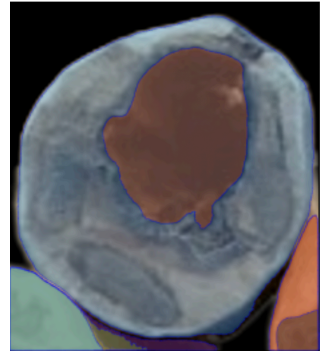 Poor

(h) 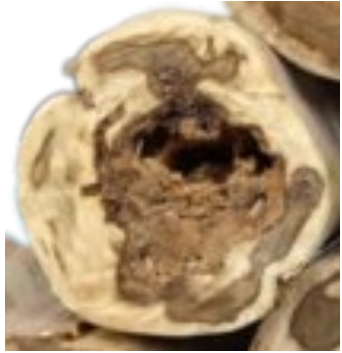 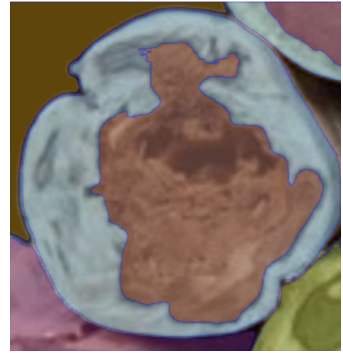 Good

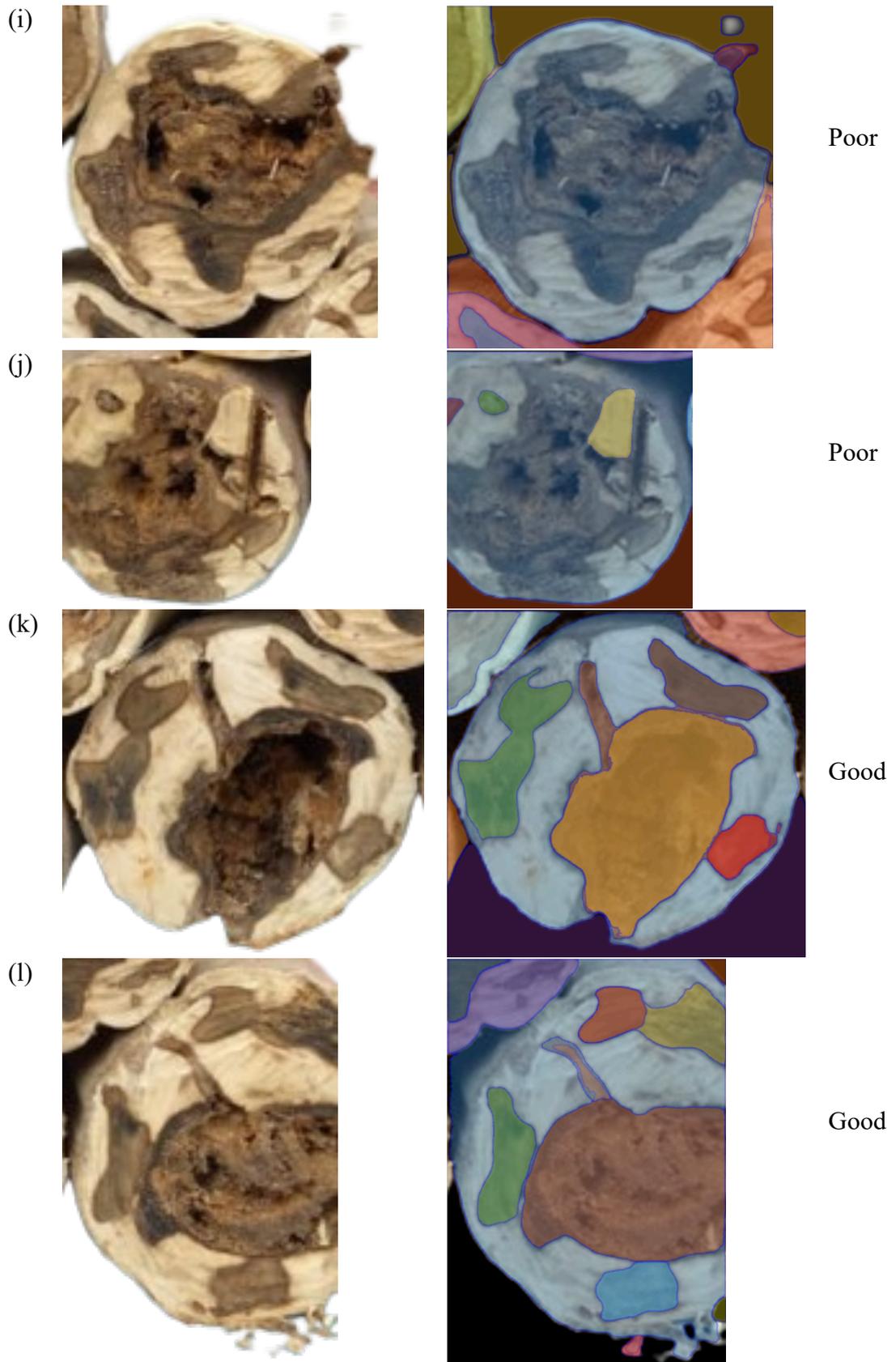

Fig. 6. SAM-produced segmentation results using a 16×16 regular grid of prompts. Based on visual inspection, we assign a class label for each segmentation result, i.e., Poor (<40.0% IoU), Moderate (40.0% - 60.0% IoU), and Good (>60.0% IoU).

Based on visual inspection alone, SAM produces a reasonable segmentation result on our dataset. Six images achieved near-perfect segmentation, i.e., Fig. 6 (b), Fig. 6 (e), Fig. 6 (f), Fig. 6 (h), Fig. 6 (k), and Fig. 6 (l). Two images, i.e., Fig. 6 (a) and Fig. 6 (g), achieved moderate segmentation, and four images achieved poor segmentation, i.e., Fig. 6 (c), Fig. 6 (d), Fig. 6 (i), and Fig. 6 (j).

The retained regions of the poorly-segmented images appear to have a lighter colour and hue, almost similar in appearance to the Removed regions. Also, their Retained regions contain a greater variety of colour and hue than the images with near-perfect segmentation. These attributes contribute to the poor segmentation result. The Retained regions belonging to the images with near-perfect segmentation appear to have a more uniform and darker colour and hue. The appearance of the Retained regions is markedly different from the Removed regions; hence it is easier to produce a good segmentation result. With a good selection of additional prompts, the region proposals will be of better quality, i.e., achieving a higher Intersection over Union (IoU) value. The prompts must capture the entire spectrum of colours and hues representing the Retained regions. Using the default 16×16 regular grid of prompts leaves the prompt selection to chance, as poor ones will produce a poor segmentation result.

Table 2: Intersection over Union (IoU).

| | Image (Rescaled) | Resolution (w x h) | IoU (%) |
|---|---|---|---|
| (a) | 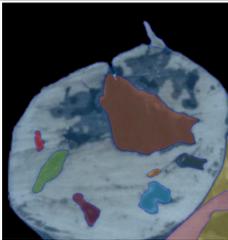 | 288 x 302 | 54.1 |
| (b) | 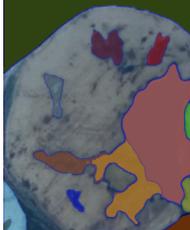 | 268 x 342 | 97.5 |
| (c) | 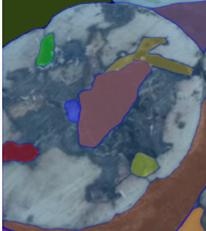 | 266 x 308 | 37.4 |
| (d) | 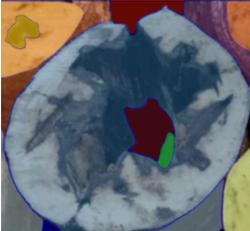 | 306 x 290 | 11.8 |
| (e) | 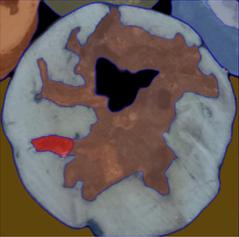 | 402 x 424 | 99.3 |
| (f) | 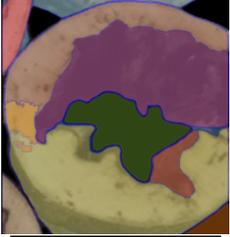 | 274 x 288 | 98.5 |
| (g) | 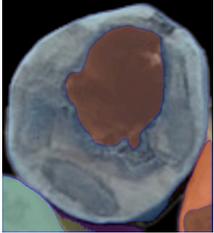 | 232 x 264 | 53.3 |
| (h) | 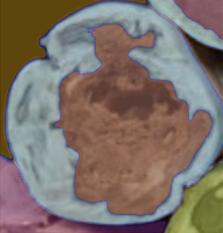 | 250 x 264 | 97.2 |
| (i) | 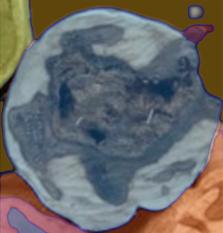 | 292 x 308 | 5.4 |
| (j) | 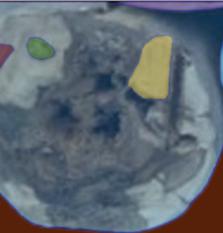 | 230 x 242 | 16.7 |
| (k) | 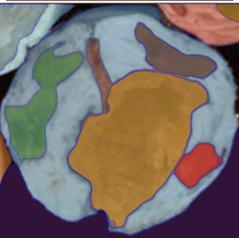 | 334 x 326 | 98.1 |
| (l) | 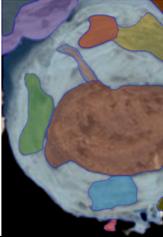 | 254 x 374 | 97.4 |

The six images with near-perfect segmentation achieved an IoU value ranging from 97.2% to 99.3%. The four images with poor segmentation achieved an IoU value ranging from 5.4% to 37.4%. Two images manage a moderate segmentation, with an IoU value of 53.3% and 54.1%. Table 3 shows each category's minimum, median, average and maximum IoU values.

Table 3: Minimum, Median, Average and Maximum IoU values.

| Segmentation. Quality | Min (%) | Med (%) | Average (%) | Max (%) |
|---|---|---|---|---|
| Good | 97.2 | 97.8 | 98.0 | 99.3 |
| Moderate | 53.3 | 53.7 | 53.7 | 54.1 |
| Poor | 5.4 | 14.3 | 17.8 | 37.4 |

As mentioned before this, the Retained regions comprise the Resinous compound and Decayed core regions. At this stage of our work, we do not implement any processing to differentiate between the two regions.

## 5 Conclusion and Future Work

Based on our initial results, we can observe the value of integrating a Deep learning segmentation algorithm inside the Agarwood resin extraction pipeline. Using default SAM, with just the regular 16×16 grid of prompts, $\frac{6}{12}$ images managed a near-perfect segmentation, each with a high IoU value. The median IoU value for these six images is $\frac{97.5+98.1}{2} = 97.8\%$. Evidently, SAM is able to differentiate between the resinous and non-resinous regions, provided that there is a high contrast of colours and/or hues between the two. SAM returns a poor segmentation result if the two regions are visually-similar.

SAM segmentation can be further refined by introducing additional prompts. Currently, this step is done manually by a human expert. In a fully autonomous pipeline, prompt creation needs to be automated. The prompts must cover all possible colours and hues of the resin compound region inside the cross-section image.

We plan to train an image-based classifier to automate the prompt creation to predict the region label. This requires extensive training data that cover all possible colours and hues of the following three classes, i.e., Healthy region, Agarwood resin compound, and Decayed core region. Since multiple unique instances (i.e., multiple combinations of colours and hues) are possible for each class label, a multi-voter scheme implemented in Hipiny and Mayol-Cuevas (2012) would be an ideal solution. Classification of an unknown small region surrounding a candidate prompt may return multi-class labels; hence a similar scheme would assist in revealing the top-voted match. The classifier can be represented using Bag-of-Words, as implemented in Hipiny and Mayol-Cuevas (2012), or as strings, as implemented in Hipiny et al. (2023). Both allow an unknown region to be matched to a labelled example from the dataset.